\documentclass[11pt]{article}
\usepackage{acl-hlt2011}
\usepackage{times}
\usepackage{latexsym}
\usepackage{amsmath}
\usepackage{multirow}
\usepackage{url}
\usepackage{graphics}
\usepackage{graphicx}
\newenvironment{myitemize}{
\vspace{-0.6\baselineskip}
\begin{itemize}
\setlength{\topsep}{0pt}
\setlength{\itemsep}{3pt}
\setlength{\parskip}{0pt}
\setlength{\parsep}{0pt}
\setlength{\partopsep}{0pt}
}{
\end{itemize}
\vspace{-0.4\baselineskip}}
\newenvironment{myenumerate}{
\vspace{-0.3\baselineskip}
\begin{enumerate}
\setlength{\topsep}{0pt}
\setlength{\itemsep}{2pt}
\setlength{\parskip}{0pt}
\setlength{\parsep}{0pt}
\setlength{\partopsep}{0pt}
}{
\end{enumerate}
\vspace{-0.2\baselineskip}}

 \title{From Once Upon a Time to Happily Ever After:\\ Tracking Emotions in Novels and Fairy Tales}

 \author{Saif Mohammad\\
     Institute for Information Technology\\
     National Research Council Canada\\
     Ottawa, Ontario, Canada, K1A 0R6\\
     {\tt saif.mohammad@nrc-cnrc.gc.ca}
 }

\date{}

\begin{document}
\maketitle
\begin{abstract}
Today we have access to unprecedented  amounts of literary texts.
However, search still relies heavily on key words.  In this paper, we
show how sentiment analysis can be used in tandem with effective
visualizations to quantify and track emotions in both individual books
and across very large collections.  We introduce the concept of
emotion word density, and using the Brothers Grimm fairy tales as
example, we show how collections of text can be organized for better
search. Using the Google Books Corpus we show how to determine an
entity's emotion associations from co-occurring words.  Finally,  we
compare emotion words in fairy tales and novels, to show that fairy
tales have a much wider range of emotion word densities than novels.
\end{abstract}

\section{Introduction}
Literary texts, such as novels, fairy tales, fables, romances, and epics have long been channels to convey emotions, both explicitly and implicitly.
With widespread digitization of text,
we now have easy access to unprecedented  amounts of such literary texts.
Project Gutenberg provides access to 34,000 books \cite{Lebert09}.\footnote{Project Gutenberg: http://www.gutenberg.org}
Google is providing access to n-gram sequences
and their frequencies 
from more than 5.2 million digitized books, as part of the {\it Google Books Corpus (GBC)} \cite{GBC}.\footnote{{\it GBC}: http://ngrams.googlelabs.com/datasets}
However, techniques to automatically access and analyze these books still rely heavily on key word searches alone.
In this paper, we show how 
sentiment analysis can be used in tandem with effective
visualizations to quantify and track emotions in both individual books
and across very large collections.
This serves many purposes, including:
\begin{myenumerate}
\item {\it Search}: Allowing search based on emotions. For example,
retrieving the darkest of the Brothers Grimm fairy tales, or finding snippets from the Sherlock Holmes series
that build the highest sense of anticipation and suspense.
\item {\it Social Analysis}: Identifying how books have portrayed different people and entities over time.
For example, what is the distribution of emotion words used in proximity to mentions of  women, race, and homosexuals.
(Similar to how Michel et al.\@ \shortcite{Michel11} tracked fame by analyzing mentions in the Google Books Corpus.)
\item {\it Comparative analysis of literary works, genres, and writing styles}: For example, is the distribution of emotion words in fairy tales significantly different from that in novels?
Do women authors use a different distribution of emotion words than their male counterparts?
Did Hans C.\@ Andersen use emotion words differently than Beatrix Potter?
\item {\it Summarization}: For example, automatically generating summaries that capture the different emotional states of the characters in a novel.
\item {\it Analyzing Persuasion Tactics}: Analyzing emotion words and their role in persuasion \cite{Mannix92,bales97}. 
\end{myenumerate}
\noindent In this paper, 
we describe how we use a large word--emotion association lexicon (described in Section 3.1)
to create a simple emotion analyzer (Section 3.2).
We present a number of visualizations that help track and analyze the use of
emotion words in individual texts and across very large collections, which is especially
useful in Applications 1, 2, and 3 described above (Section 4).
We introduce the concept of emotion word density, and using the Brothers Grimm fairy tales as an example, we show how collections of text can be organized
for better search (Section 5). Using the Google Books Corpus we show how to determine emotion associations portrayed in books towards different entities (Section 6). 
Finally, for the first time, we compare a collection of novels and a collection of fairy tales using an emotion lexicon to show that
fairy tales have a much wider distribution of emotion word densities than novels.

The emotion analyzer recognizes words with positive polarity (expressing a favorable sentiment towards an entity), negative polarity (expressing an unfavorable sentiment towards an entity), and no polarity (neutral). 
It also  associates words with joy, sadness, anger, fear,
trust, disgust, surprise, anticipation, which are argued to be the eight
basic and prototypical emotions \cite{Plutchik80}. 

This work is part of a broader project to provide an affect-based interface to Project Gutenberg.
Given a search query, the goal is to provide users with relevant plots presented in this paper,
as well as  ability to search for text snippets from multiple texts that have high
emotion word densities. 


  \section{Related work}
Over the last decade, there has been considerable work in sentiment analysis, especially in determining whether a term has a positive
or negative  polarity \cite{Lehrer74,Turney03,MohammadDD09}. 
There is also work in more sophisticated aspects of sentiment, for example, in detecting emotions such as
anger, joy, sadness, fear, surprise, and disgust \cite{Bellegarda10,MohammadT10,AlmRS05,AlmRS05}.
The technology is still developing and it can be unpredictable when dealing with short sentences,
but it has been shown to be reliable when drawing conclusions from large amounts of text \cite{DoddsD10,PangL08}.

Automatic analysis of emotions in text has so far had to rely on small emotion lexicons. 
The WordNet Affect Lexicon (WAL) \cite{StrapparavaV04} has a few hundred words
annotated with associations to a number of affect categories including the six Ekman emotions (joy, sadness, anger, fear,
disgust, and surprise).\footnote{WAL: http://wndomains.fbk.eu/wnaffect.html} 
 General Inquirer (GI) \cite{Stone66} has 11,788 words labeled with 182
categories of word tags, including positive and negative polarity.\footnote{GI: http://www.wjh.harvard.edu/$\sim$inquirer} 
 We use the NRC Emotion Lexicon \cite{MohammadY11,MohammadT10}, a large set of human-provided word--emotion association ratings, in our experiments.\footnote{Please send an e-mail to
 saif.mohammad@nrc-cnrc.gc.ca to obtain the latest version of the NRC Emotion Lexicon.}

Empirical assessment of emotions in literary texts has sometimes relied on human annotation of the texts,
but this has restricted the number of texts analyzed. For example, Alm and Sproat \shortcite{Alm05} annotated
22 Brothers Grimm fairy tales to show that fairy tales often began with a neutral sentence and
ended with a happy sentence. Here we use out-of-context word--emotion associations
and analyze individual texts to very large collections. We rely on information from many words
to provide a strong enough signal to overcome individual errors due to out-of-context annotations.

\section{Emotion Analysis}

\subsection{Emotion Lexicon}

The NRC Emotion Lexicon was created by crowdsourcing to Amazon's Mechanical Turk, and it is described in \cite{MohammadY11,MohammadT10};
we briefly summarize below.

The 1911 {\it Roget Thesaurus} was used as the source for target terms.\footnote{Roget's Thesaurus: www.gutenberg.org/ebooks/10681}  
Only those thesaurus words that occurred more than 120,000 times in the Google n-gram corpus were annotated for version 0.92 of the lexicon
which we use for the experiments described in this paper.\footnote{The Google N-gram Corpus is available through the Linguistic Data Consortium.}

 The {\it Roget's Thesaurus} groups related words into about a thousand categories,
 which can be thought of as coarse senses or concepts \cite{Yarowsky92}.
 If a word is ambiguous, then it is listed in more than one category.
 Since a word may have different emotion associations when used in different senses,
 word-sense level annotations were obtained by first asking 
an automatically
generated word-choice question pertaining to the target:

\vspace*{1mm}
{\small
\noindent Q1. Which word is closest in meaning  
to {\it shark} (target)?

 \begin{minipage}[t]{4mm}
 \end{minipage}
 \begin{minipage}[t]{1.7cm}
  \begin{myitemize}
 \item {\it car}
 \end{myitemize}
 \end{minipage}
 \begin{minipage}[t]{1.9cm}
  \begin{myitemize}
 \item {\it tree}
 \end{myitemize}
 \end{minipage}
 \begin{minipage}[t]{1.7cm}
  \begin{myitemize}
 \item {\it fish}
 \end{myitemize}
 \end{minipage}
 \begin{minipage}[t]{1.6cm}
  \begin{myitemize}
 \item {\it olive}
 \end{myitemize}
 \end{minipage}
}

\noindent The near-synonym for Q1 is taken from the thesaurus, and the distractors are randomly chosen words.
This question guides the annotator to the desired sense of the target word.
It is followed by ten questions asking if the target  is associated with positive sentiment, negative sentiment,
 anger, fear, joy, sadness, disgust, surprise, trust, and anticipation. The questions were phrased exactly
  as described in Mohammad and Turney \shortcite{MohammadT10}.

If an annotator answers Q1 incorrectly,
then information obtained from the remaining questions is discarded.
Thus, even though there were no gold standard correct answers to the emotion association questions, likely incorrect annotations were filtered out.
About 10\% of the annotations were discarded because of an incorrect response to Q1.

Each term was annotated by 5 different people.
For 74.4\% of the 
instances, all five annotators agreed on whether a term is associated
with a particular emotion or not. For 16.9\% of the instances four out of five people agreed with each other.
The information from multiple annotators for a particular term was combined
by taking the majority vote. 
The lexicon has entries for about 24,200 word--sense pairs. The information from
different senses of a word was combined
by taking the union of all emotions associated with the different senses of the word.
This resulted in a word-level emotion association lexicon for about 14,200 word types.

\subsection{Text Analysis}
Given a target text, the system determines which of the words exist in our emotion lexicon
and calculates ratios such as the number of words associated with an emotion to the total number of emotion words 
in the text. This simple approach may not be reliable in determining if a particular
sentence is expressing a certain emotion, but it is reliable in determining if a large piece of text
has more emotional expressions compared to others in a corpus.
Example applications include 
clustering literary texts based on the distributions of emotion words,
analyzing gender-differences in email \cite{MohammadY11},
and detecting spikes in anger words in close proximity to
 mentions of a target product in a twitter stream \cite{DiazR02,DubeM96}.

  \begin{figure}[t]
  \begin{center}
  \includegraphics[width=\columnwidth]{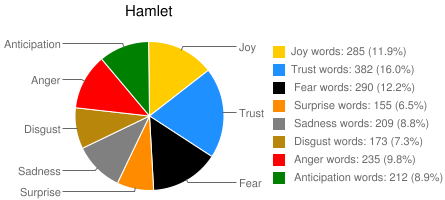}
  \end{center}
  \vspace*{-3mm}
  \caption{{\bf Emotions pie chart} of Shakespeare's tragedy {\it Hamlet}. (Text from Project Gutenberg.)}
  \vspace*{-1mm}
  \label{fig:hamlet}
  \end{figure}

  \begin{figure}[t]
  \begin{center}
  \includegraphics[width=\columnwidth]{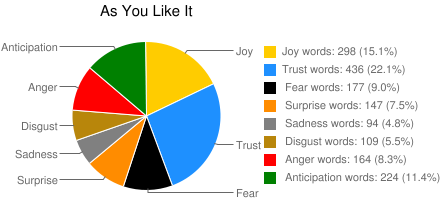}
  \end{center}
  \vspace*{-3mm}
  \caption{{\bf Emotions pie chart} of Shakespeare's comedy {\it As you like it}. (Text from Project Gutenberg.)}
  \vspace*{2mm}
  \label{fig:as you like it}
  \end{figure}

  \begin{figure}[t!]
  \begin{center}
  \includegraphics[width=\columnwidth]{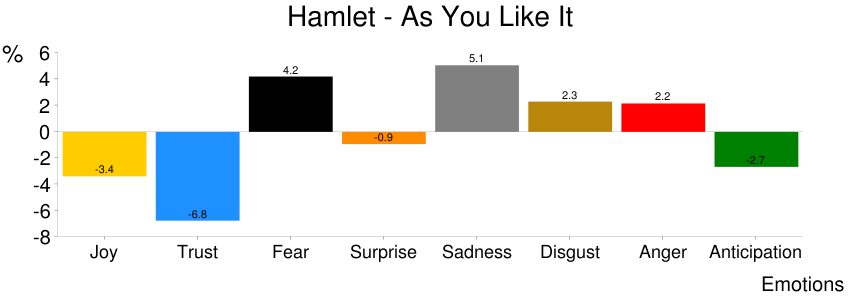}
  \end{center} 
  \vspace*{-3mm}
  \caption{Difference in percentage scores for each of the eight basic emotions in {\it Hamlet} and {\it As you like it}.}
  \vspace*{-3mm}
  \label{fig:hamlet-asyoulikeit-deviation}
  \end{figure}

\section{Visualizations of Emotions}

\subsection{Distributions of Emotion Words}

Figures \ref{fig:hamlet} and \ref{fig:as you like it} show the percentages of emotion words 
in Shakespeare's famous tragedy, {\it Hamlet}, and his comedy, {\it As you like it}, respectively.
Figure \ref{fig:hamlet-asyoulikeit-deviation} conveys the difference between the two novels even more explicitly
by showing only the difference in percentage scores for each of the emotions.
Emotions are represented by colours as per a study on word--colour associations \cite{Mohammad11a}.


  \begin{figure}[t]
  \begin{center}
  \includegraphics[width=\columnwidth]{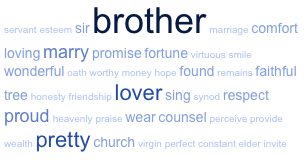}
  \end{center}
  \vspace*{-5mm}
  \caption{{\it Hamlet - As You Like It}: relative-salience word cloud for trust words.}
  \label{fig:hamlet-asyoulikeit-wordcloud-trust}
  \end{figure}


  \begin{figure}[t!]
  \begin{center}
  \includegraphics[width=\columnwidth]{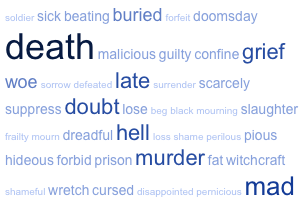}
  \end{center}
  \vspace*{-5mm}
  \caption{{\it Hamlet - As You Like It}: relative-salience word cloud for sadness words.}
  \vspace*{-3mm}
  \label{fig:hamlet-asyoulikeit-wordcloud-sadness}
  \end{figure}

Observe how one can clearly see that {\it Hamlet} has more fear, sadness, disgust, and anger,
and less joy, trust, and anticipation.
The bar graph is effective at conveying the extent to which an emotion is more prominent in one
text than another, but it does not convey the source of these emotions.
Therefore, 
we calculate the {\it relative salience} of an emotion word $w$ 
across two target texts $T_1$ and $T_2$:
\vspace*{-2mm}
\begin{equation}
\label{eq:salience}
\text{RelativeSalience}(w|T_1,T_2) = \frac{f_1}{N_1} - \frac{f_2}{N_2} 
\end{equation}
\noindent Where, $f_1$ and $f_2$ are the frequencies of $w$ in $T_1$ and $T_2$, respectively.
$N_1$ and $N_2$ are the total number of word tokens in $T_1$ and $T_2$.
Figures \ref{fig:hamlet-asyoulikeit-wordcloud-trust} and \ref{fig:hamlet-asyoulikeit-wordcloud-sadness} depict 
snippets of relative-salience word clouds of trust words 
and sadness words across {\it Hamlet} and {\it As You Like it}.
Our emotion analyzer uses Google's freely available software 
to create word clouds.\footnote{Google word cloud: http://visapi-gadgets.googlecode.com/\\svn/trunk/wordcloud/doc.html}

\subsection{Flow of Emotions}

\begin{figure}[t]
  \begin{center}
  \includegraphics[width=\columnwidth]{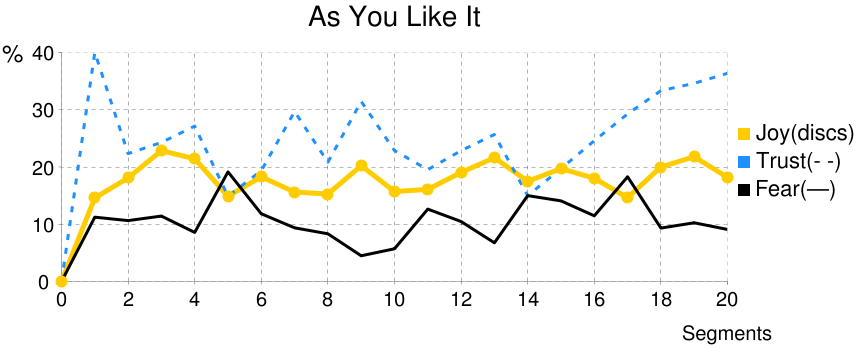}
  \end{center}
  \vspace*{-5mm}
  \caption{Timeline of the emotions in {\it As You Like It}.} 
  \vspace*{-3mm}
  \label{fig:asyoulikeit-timeline}
  \end{figure}

\begin{figure}[t!]
  \begin{center}
  \includegraphics[width=\columnwidth]{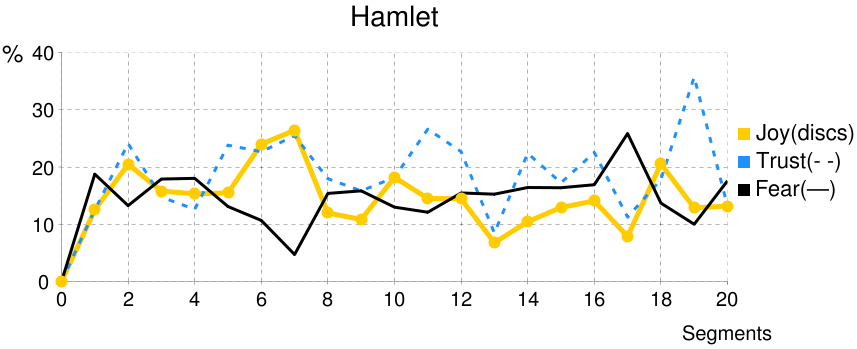}
  \end{center}
  \vspace*{-5mm}
  \caption{Timeline of the emotions in {\it Hamlet}.} 
  \vspace*{-3mm}
  \label{fig:hamlet-timeline}
  \end{figure}

\begin{figure}[t!]
  \begin{center}
  \includegraphics[width=\columnwidth]{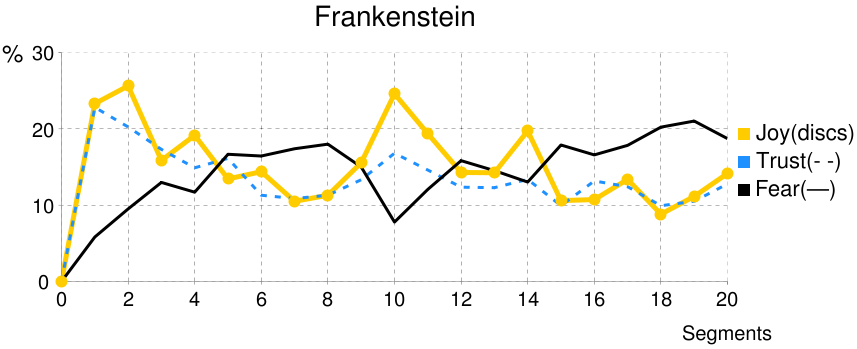}
  \end{center}
  \vspace*{-5mm}
  \caption{Timeline of the emotions in {\it Frankenstein}.} 
  \label{fig:frankenstein-timeline}
  \end{figure}

Literary researchers as well as casual readers may be interested in noting how
the use of emotion words has varied through the course of a book.
Figure \ref{fig:asyoulikeit-timeline}, \ref{fig:hamlet-timeline}, and \ref{fig:frankenstein-timeline} show the flow of joy, trust, and fear
in {\it As You Like it} (comedy), {\it Hamlet} (tragedy), and {\it Frankenstein} (horror), respectively.
As expected, the visualizations depict the novels to be progressively more dark than the previous ones in the list. Also
that {\it Frankenstein} is much darker in the final chapters.

\section{Emotion Word Density}

  \begin{figure*}[t]
  \begin{center}
  \includegraphics[width=2\columnwidth]{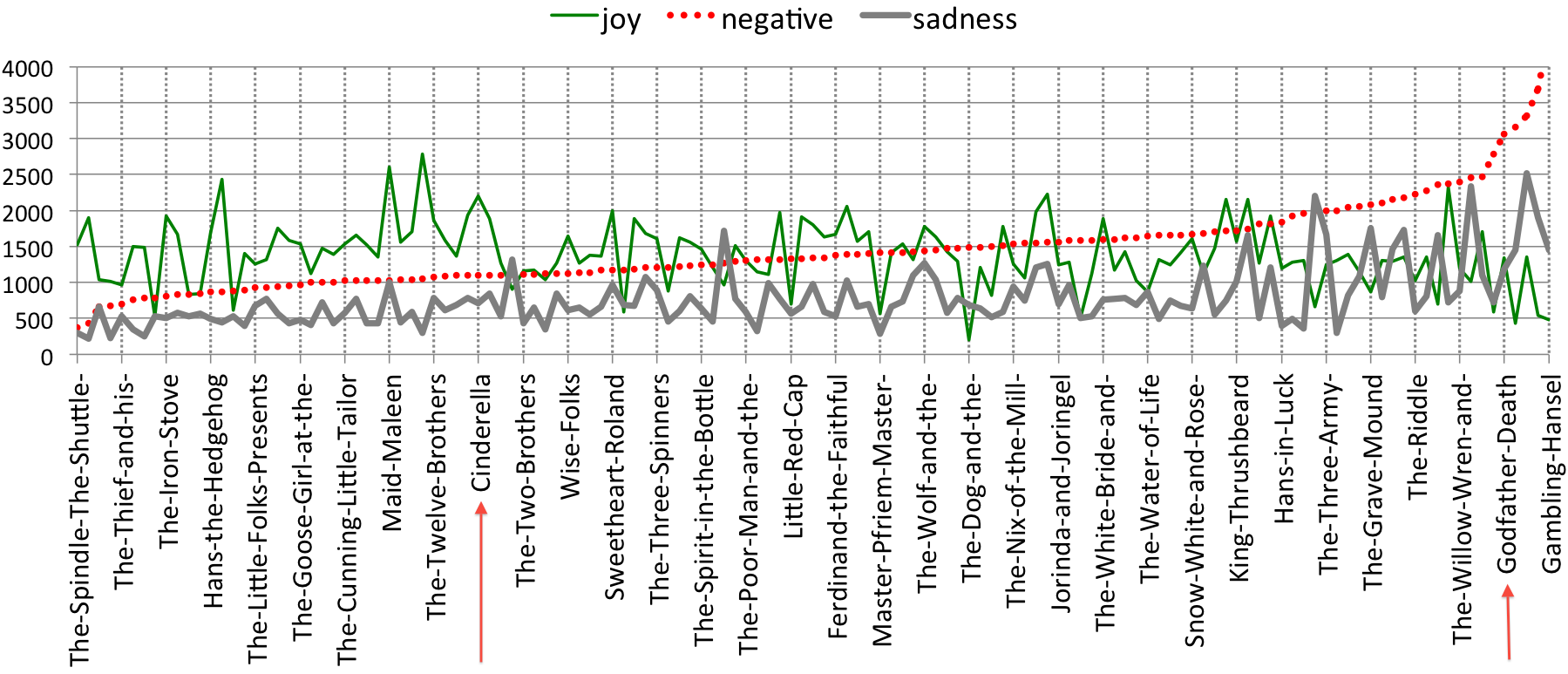}
  \end{center}
  \vspace*{-3mm}
  \caption{The Brothers Grimm fairy tales arranged in increasing order of negative word density (number of negative words in every 10,000 words). 
  The plot is of 192 stories but the x-axis has labels for only a few due to lack of space. 
A user may select any two tales, say {\it Cinderella} and {\it Godfather Death} (follow arrows), to reveal
Figure 10.}
  \label{fig:grimm-plot}
  \end{figure*}

\begin{figure}[t]
  \vspace*{-2mm}
  \begin{center}
  \includegraphics[width=\columnwidth]{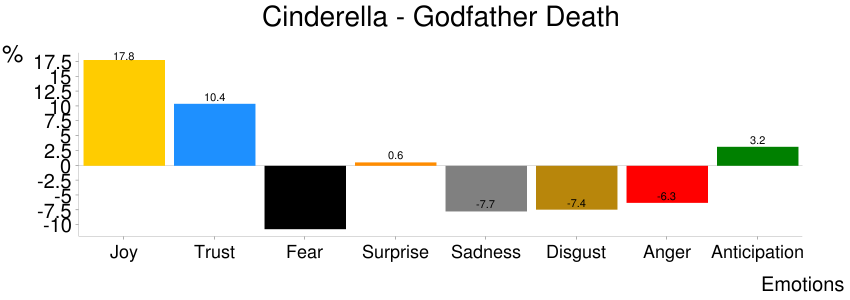}
  \end{center}
  \vspace*{-5mm}
  \caption{The difference in percentages of emotion words across {\it Cinderella} and {\it Godfather Death}.}
  \vspace*{-4mm}
  \label{fig:Cinderella-Godfather-dev-bar}
  \end{figure}

\begin{figure}[t]
  \begin{center}
  \includegraphics[width=\columnwidth]{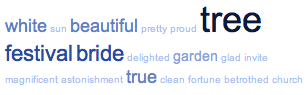}
  \end{center}
  \vspace*{-4mm}
  \caption{{\it Cinderella - Godfather Death}: Relative salience word cloud of joy.}
  \vspace*{-4mm}
  \label{fig:Cinderella-Godfather-Death-wordcloud-joy}
  \end{figure}

Apart from determining the relative percentage of different words, the use of emotion words in a book can also be quantified
by calculating the number of emotion words one is expected to see on reading every $X$ words. We will refer to this
metric as {\it emotion word density}. All emotion densities reported in this paper are for $X=10,000$.
The dotted line in Figure \ref{fig:grimm-plot} shows the negative word density plot of 192 fairy tales collected by Brothers Grimm.
The joy and sadness word densities are also shown---the thin and thick lines, respectively.
A person interested in understanding the use of emotion words in the fairy tales collected by Brothers Grimm
can further select any two fairy tales from the plot, to reveal a bar graph showing the difference in percentages
of emotions in the two texts.
Figure \ref{fig:Cinderella-Godfather-dev-bar} shows the difference bar graph of {\it Cinderella} and {\it Godfather Death}.
Figures \ref{fig:Cinderella-Godfather-Death-wordcloud-joy} depicts
the relative-salience word cloud of joy words
 across the two fairy tales. The relative-salience word cloud of fear included: {\it death, ill, beware, poverty, devil, astray, risk, illness,
threatening, horrified} and {\it revenge}.

\section{Emotions Associated with Targets}
Words found in proximity of target entities can be good indicators of emotions associated with the targets.
Google has released n-gram frequency data from all the books they have scanned up to July 15, 2009.\footnote{Google books data: http://ngrams.googlelabs.com/datasets.}
The data consists of 5-grams along with the number of times they were used in books published in every year from 1600 to 2009.
We analyzed the 5-gram files (about 800GB of data) to quantify the emotions associated with different target
entities. We ignored data from books published before 1800 as that period is less comprehensively covered by Google books.
We chose to group the data into five-year bins, though other groupings are reasonable as well.
Given a target entity of interest, the system identifies all 5-grams that contain the target word,
identifies all the emotion words in those n-grams (other than the target word itself), and calculates percentages of emotion words.

  \begin{figure*}[t!]
  \begin{center}
  \includegraphics[width=1.8\columnwidth]{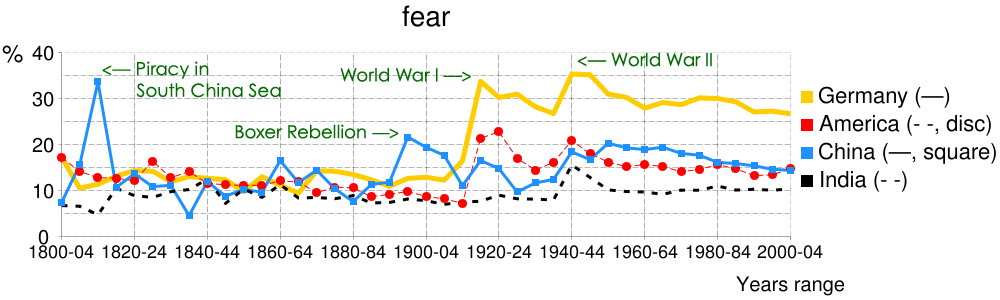}
  \end{center}
  \vspace*{-6mm}
  \caption{Percentage of {\bf fear} words in close proximity to occurrences of {\it America, China, Germany,} and {\it India} in books from the year 1800 to 2004.
Source: 5-gram data released by Google.}
  \vspace*{-2mm}
  \label{fig:countries-timeline}
  \end{figure*}

  \begin{figure}[t]
  \begin{center}
  \includegraphics[width=\columnwidth]{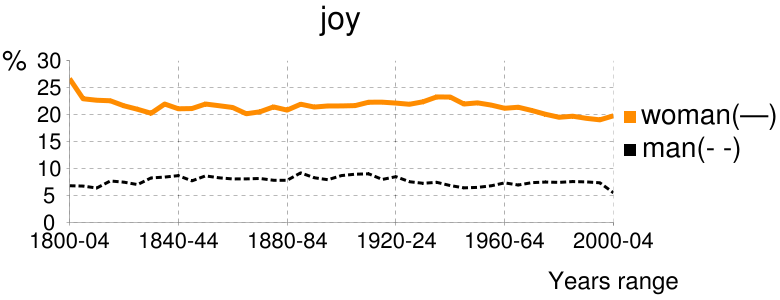}
  \end{center}
  \vspace*{-5mm}
  \caption{Percentage of {\bf joy} words in close proximity to occurrences of {\it man} and {\it woman} in books.}
  \vspace*{-4mm}
  \label{fig:manwoman-joy-timeline}
  \end{figure}

Figure \ref{fig:countries-timeline} shows the percentage of fear words in the n-grams of different countries.
Observe, that there is a marked rise of fear words around World War I (1914--1918) for Germany, America, and China.
There is a spike for China around 1900, likely due to the unrest leading up to the Boxer Rebellion (1898--1901).\footnote{http://en.wikipedia.org/wiki/Boxer\_Rebellion}
The 1810--1814 spike for China is probably correlated with descriptions of piracy in the South China Seas, since the era of the commoner-pirates of mid-Qing dynasty came to an end in 
1810.\footnote{http://www.iias.nl/nl/36/IIAS\_NL36\_07.pdf}
India does not see a spike during World War I, but has a spike in the 1940's
probably reflecting heightened vigor in the independence struggle (Quit India Movement of 1942\footnote{http://en.wikipedia.org/wiki/Quit\_India\_Movement}) and growing involvement in 
World War II (1939--1945).\footnote{http://en.wikipedia.org/wiki/India\_in\_World\_War\_II}

\begin{figure}[t]
  \begin{center}
  \includegraphics[width=\columnwidth]{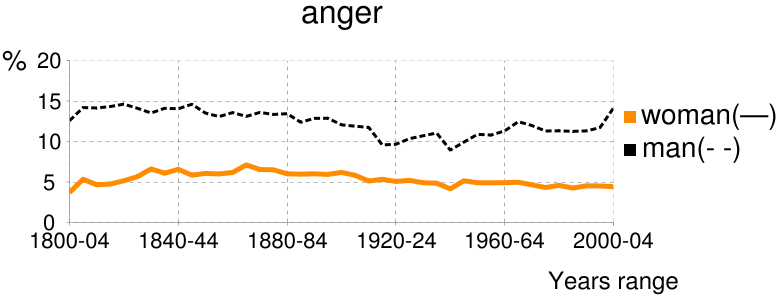}
  \end{center}
  \vspace*{-5mm}
  \caption{Percentage of {\bf anger} words in close proximity to occurrences of {\it man} and {\it woman} in books.} 
  \vspace*{-4mm}
  \label{fig:manwoman-anger-timeline}
  \end{figure}

 Figures \ref{fig:manwoman-joy-timeline} shows two curves for the percentages of joy words in 5-grams that include {\it woman} and {\it man}, respectively.
 Figures \ref{fig:manwoman-anger-timeline} shows similar curves for anger words.

\section{Emotion Words in Novels vs. Fairy Tales}
Novels and fairy tales are two popular forms of literary prose. Both forms tell a story,
but a fairy tale has certain distinct characteristics such as (a) archetypal characters (peasant, king)
(b) clear identification of good and bad characters, (c) happy ending, (d) presence of magic and magical creatures, and (d) a clear moral \cite{Jones02}.
Fairy tales are extremely popular and appeal to audiences through emotions---they
convey personal concerns, subliminal fears, wishes, and fantasies in an exaggerated manner \cite{Kast93,Jones02,Orenstein03}.
However, there have not been any large-scale empirical studies to compare affect 
in fairy tales and novels. Here for the first time, we compare the use of emotion-associated  words in fairy tales and novels
using a large lexicon.

Specifically, we are interested in determining whether: (1) fairy tales on average have a higher emotional density
than novels, (2) different fairy tales focus on different emotions such that some fairy tales have high densities for certain
emotion, whereas others have low emotional densities for those same emotions.

\begin{table*}[t!]
\centering
 \resizebox{\textwidth}{!}{
\begin{tabular}{l rrrr rrrr rr rrrr rrrr rr}
\hline
		&\multicolumn{2}{c}{\bf anger}  &\multicolumn{2}{c}{\bf anticip.}       &\multicolumn{2}{c}{\bf disgust}    &\multicolumn{2}{c}{\bf fear}   &\multicolumn{2}{c}{\bf joy}       &\multicolumn{2}{c}{\bf sadness}    &\multicolumn{2}{c}{\bf surprise}   &\multicolumn{2}{c}{\bf trust}\\
&mean  &$\sigma$	&mean  &$\sigma$	&mean  &$\sigma$	&mean  &$\sigma$	&mean  &$\sigma$	&mean  &$\sigma$	&mean  &$\sigma$	&mean  &$\sigma$	\\
\hline
CEN		&746 &162   &1230 &126  &591 &135   &975 &225   &1164 &196  &785 &159   &628 &93    &1473 &190\\
FTC    &749 &393    &1394 &460 	&682 &460 	&910 &454 	&1417 &467  &814 &443 	&680 &325 	&1348 &491\\
\hline
\end{tabular}
}
\vspace*{-4mm}
\caption{Density of emotion words in novels and fairy tales: number of emotion words in every 10,000 words.}
\label{tab:emo density}
 \vspace*{-1mm}
\end{table*}

We used the Corpus of English Novels (CEN) and the Fairy Tale Corpus (FTC) for our experiments.\footnote{CEN: https://perswww.kuleuven.be/$\sim$u0044428/cen.htm\\
FTC: https://www.l2f.inesc-id.pt/wiki/index.php/Fairy\_tale\_corpus}
The Corpus of English Novels is a collection of 292 novels written between 1881 and 1922 by 25 British and American novelists.
It was compiled from Project Gutenberg at the Catholic University of Leuven by Hendrik de Smet. 
It consists of about 26 million words.
The Fairy Tale Corpus \cite{Lobo10} has 453 stories, close to 1 million words, downloaded from Project Gutenberg.
Even though many  fairy tales have a strong oral tradition, the stories in this collection were compiled,
translated, or penned in the 19th century by the Brothers Grimm, Beatrix Potter, and Hans C.\@ Andersen to name a few.

We calculated the polarity and emotion word density of each of the novels in CEN and each of the fairy tales in FTC.
Table 1 lists the mean densities as well as standard deviation for each of the eight basic emotions in the two corpora.
We find that the mean densities for anger and sadness across CEN and FTC are not significantly different. 
However,
fairy tales have significantly higher anticipation, disgust, joy, and surprise densities when compared to novels ($p < 0.001$).
On the other hand, they have significantly lower trust word density than novels.
Further, the standard deviations for all eight emotions are significantly different across the two corpora ($p < 0.001$).
The fairy tales, in general, have a much larger standard deviation than the novels. 
Thus for each of the 8 emotions, there are more fairy tales than novels
having high emotion densities and there are more fairy tales than novels having low emotion densities.

Table \ref{tab:pol density} lists the mean densities as well as standard deviation for negative and positive polarity words in the two corpora.
The table states, for example, that for every 10,000 words in the CEN, one can expect to see about 1670 negative words.
We find that fairy tales, on average, have a significantly lower number of negative terms, and a significantly
higher number of positive words ($p < 0.001$). 

\begin{figure}[t!]
  \begin{center}
  \includegraphics[width=\columnwidth]{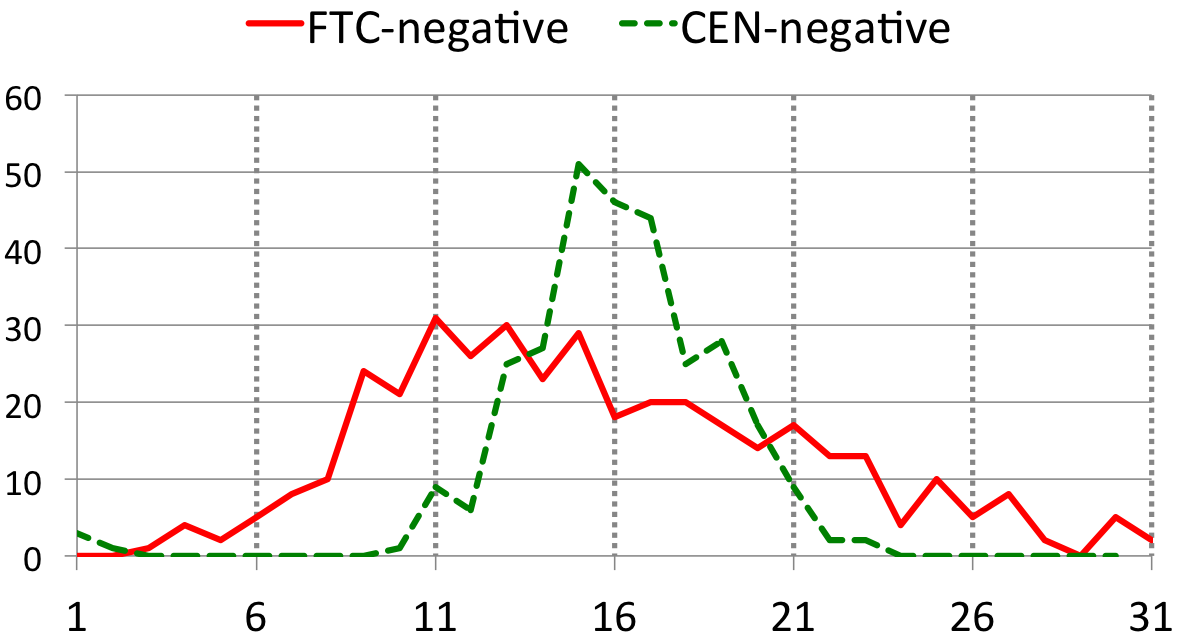}
  \end{center}
  \caption{Histogram of texts with different negative word densities.  On the x-axis: 1 refers to density between 0 and 100, 2 refers to 100 to 200, and so on.}
  \label{fig:neg-diff}
  \end{figure}

\begin{figure}[t!]
 \begin{center}
 \includegraphics[width=\columnwidth]{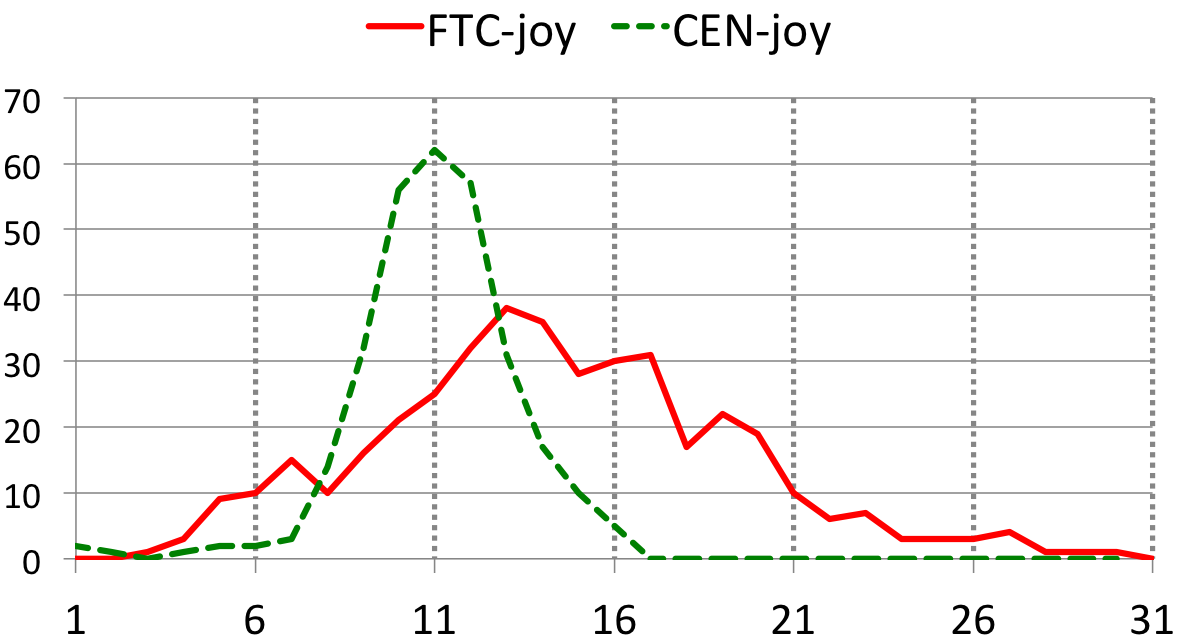}
 \end{center}
 \caption{Histogram of texts with different joy word densities.} 
 \label{fig:joy-diff}
  \end{figure}

\begin{figure}[t!]
  \begin{center}
  \includegraphics[width=\columnwidth]{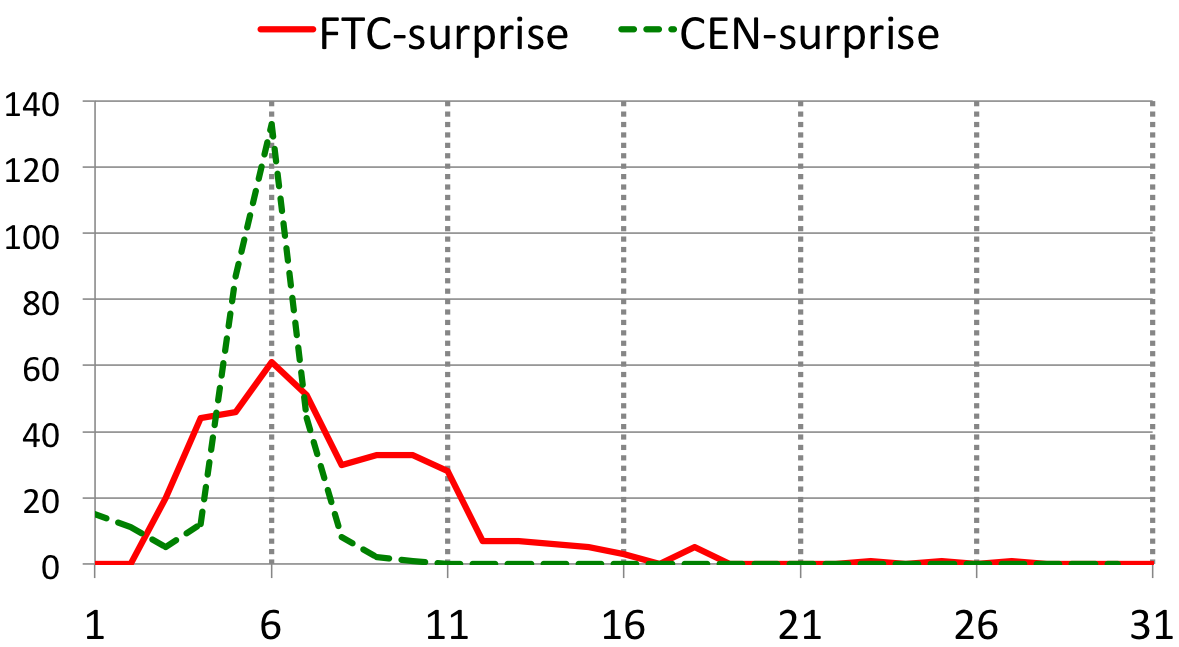}
  \end{center}
  \caption{Histogram of texts with different surprise word densities.} 
  \label{fig:surprise-diff}
  \end{figure}

\begin{figure}[t!]
  \begin{center}
  \includegraphics[width=\columnwidth]{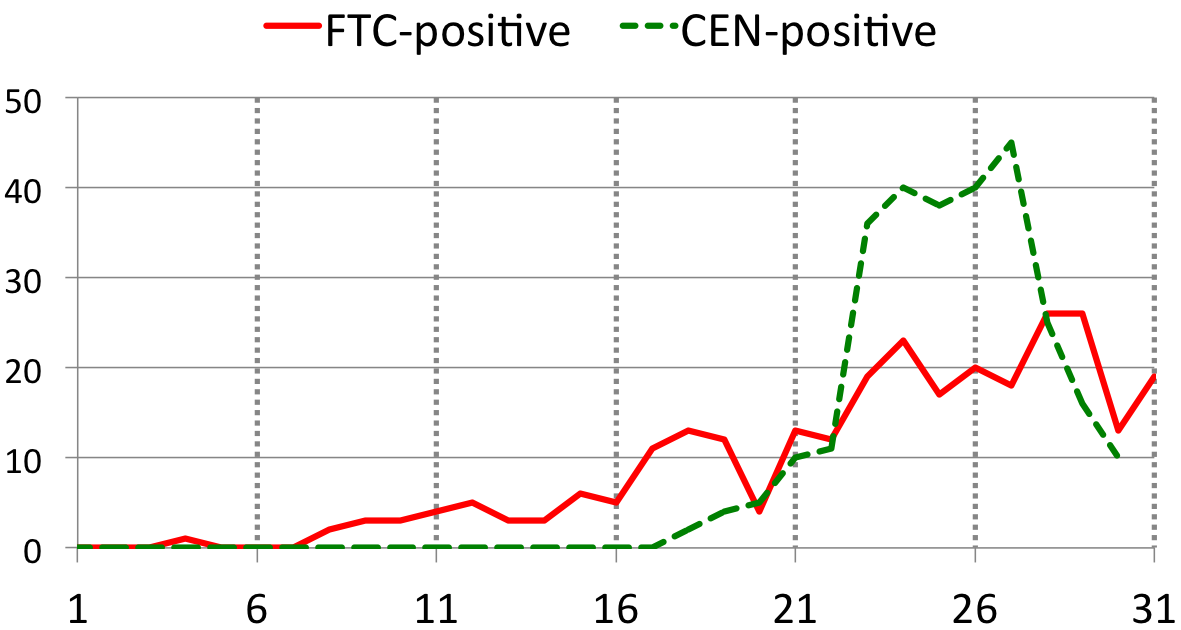}
  \end{center}
  \caption{Histogram of texts with different positive word densities. On the x-axis: 1 refers to density between 0 and 100, 2 refers to 100 to 200, and so on.}
  \label{fig:pos-diff}
  \end{figure}

\begin{figure}[t!]
  \begin{center}
  \includegraphics[width=\columnwidth]{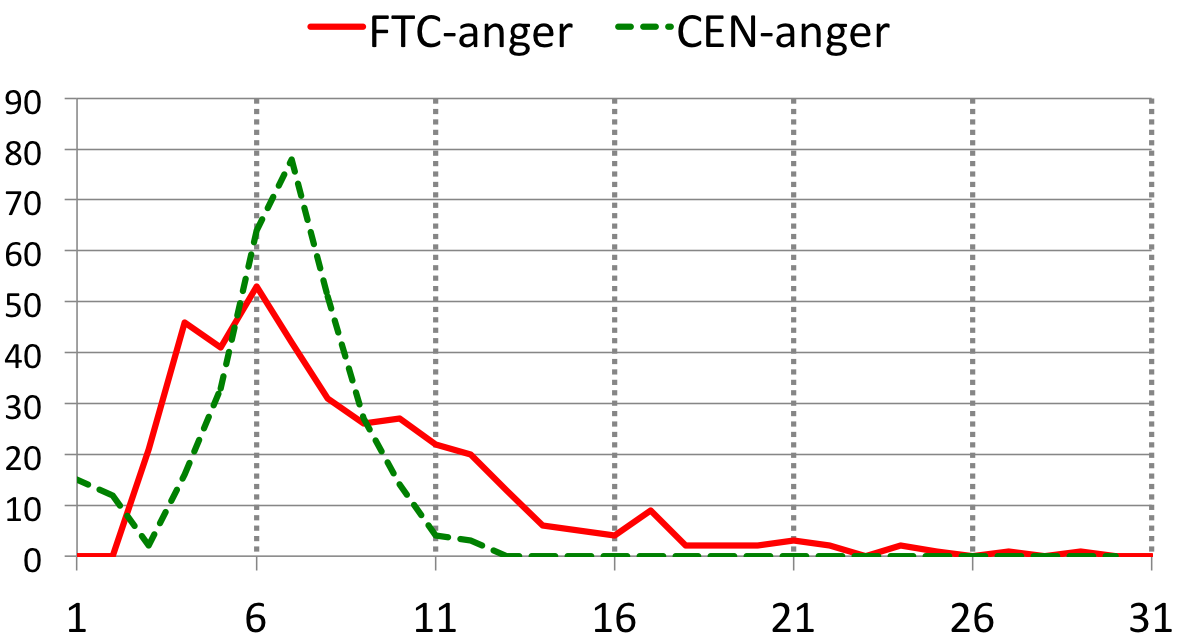}
  \end{center}
  \caption{Histogram of texts with different anger word densities.} 
  \label{fig:anger-diff}
  \end{figure}

\begin{figure}[t!]
  \begin{center}
  \includegraphics[width=\columnwidth]{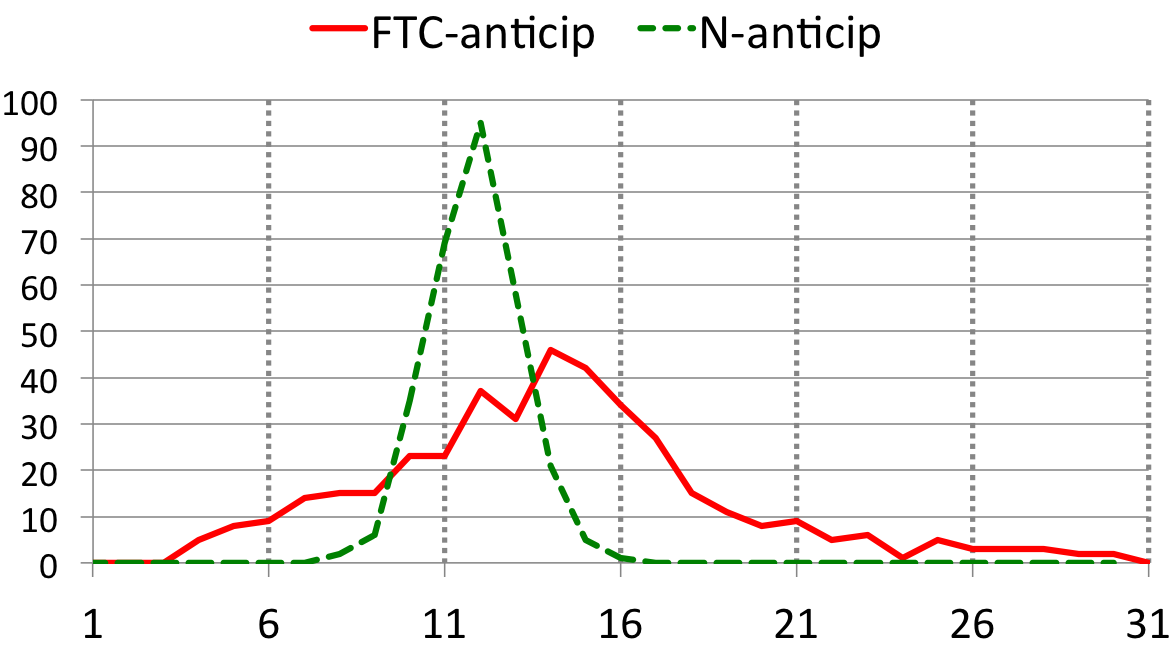}
  \end{center}
  \caption{Histogram of texts with different anticip word densities.} 
  \label{fig:anticip-diff}
  \end{figure}

In order to obtain a better sense of the distribution of emotion densities, we generated histograms by counting all texts that had emotion densities between 0--99, 100--199, 200--399, and so on.
A large standard deviation for fairy tales could be due to one of at least two reasons: 
(1) the histogram has a bimodal distribution---most of the fairy tales have extreme emotion densities (either much higher than that of the novels, or much smaller).
(2) the histogram approaches a normal distribution such that more fairy tales than novels have extreme emotion densities. 
Figures \ref{fig:neg-diff} through \ref{fig:anticip-diff} show histograms comparing novels and fairy tales for positive and negative polarities, as well as for a few emotions.
Observe that fairy tales do not have a bimodal distribution, and case (2) holds true.

\begin{table}[]
\centering
 {\small
\begin{tabular}{l rrrr}
\hline
 &\multicolumn{2}{c}{\bf negative}   &\multicolumn{2}{c}{\bf positive}\\
&mean  &$\sigma$    &mean  &$\sigma$\\
\hline
CEN		&1670 &243  &2602 &278\\
FTC 	&1543 &613    &2808 &726 \\
\hline
\end{tabular}
}
\vspace*{-2mm}
\caption{Density of polarity words in novels and fairy tales: number of polar words in every 10,000 words.}
\label{tab:pol density}
\vspace*{-3mm}
\end{table}

\section{Conclusions and Future Work}
We presented an emotion analyzer that relies on the powerful word--emotion association lexicon.
We presented a number of visualizations that help track and analyze the use of
emotion words in individual texts and across very large collections.
We introduced the concept of emotion word density, and using the Brothers Grimm fairy tales as an example, we showed how collections of text can be organized for better search. 
Using the Google Books Corpus we showed how to determine emotion associations portrayed in books towards different entities.
Finally, for the first time, we compared a collection of novels and a collection of fairy tales using the emotion lexicon to show that
fairy tales have a much wider distribution of emotion word densities than novels.

This work is part of a broader project to provide an affect-based interface to Project Gutenberg.
Given a search query, the goal is to provide users with relevant plots presented in this paper.
Further, they will be able to search for snippets from multiple texts that have strong
emotion word densities. 

 \section*{Acknowledgments}
 Grateful thanks to Peter Turney and Tara Small for many wonderful ideas.
 Thanks to Tony (Wenda) Yang for creating an online emotion analyzer.

 \clearpage

\clearpage



\clearpage

\bibliography{references}

\begin{thebibliography}{}

\bibitem[\protect\citename{Alm and Sproat}2005]{Alm05}
Cecilia~O. Alm and Richard Sproat, 2005.
\newblock {\em Emotional sequencing and development in fairy tales}, pages
  668–--674.
\newblock Springer.

\bibitem[\protect\citename{Alm \bgroup et al.\egroup }2005]{AlmRS05}
Cecilia~Ovesdotter Alm, Dan Roth, and Richard Sproat.
\newblock 2005.
\newblock Emotions from text: {M}achine learning for text-based emotion
  prediction.
\newblock In {\em Proceedings of the Joint Conference on HLT--EMNLP},
  Vancouver, Canada.

\bibitem[\protect\citename{Bales}1997]{bales97}
Richard Bales.
\newblock 1997.
\newblock {\em Persuasion in the French personal novel: Studies of
  Chateaubriand, Constant, Balzac, Nerval, and Fromentin}.
\newblock Summa Publications, Birmingham, Alabama.

\bibitem[\protect\citename{Bellegarda}2010]{Bellegarda10}
Jerome Bellegarda.
\newblock 2010.
\newblock Emotion analysis using latent affective folding and embedding.
\newblock In {\em Proceedings of the NAACL-HLT 2010 Workshop on Computational
  Approaches to Analysis and Generation of Emotion in Text}, Los Angeles,
  California.

\bibitem[\protect\citename{D\'{i}az and Ruíz}2002]{DiazR02}
Ana B.~Casado D\'{i}az and Francisco J.~M\'{a}s Ruíz.
\newblock 2002.
\newblock The consumer’s reaction to delays in service.
\newblock {\em International Journal of Service Industry Management},
  13(2):118--140.

\bibitem[\protect\citename{Dodds and Danforth}2010]{DoddsD10}
Peter Dodds and Christopher Danforth.
\newblock 2010.
\newblock Measuring the happiness of large-scale written expression: Songs,
  blogs, and presidents.
\newblock {\em Journal of Happiness Studies}, 11:441--456.
\newblock 10.1007/s10902-009-9150-9.

\bibitem[\protect\citename{Dub\'{e} and Maute}1996]{DubeM96}
Laurette Dub\'{e} and Manfred~F. Maute.
\newblock 1996.
\newblock The antecedents of brand switching, brand loyalty and verbal
  responses to service failure.
\newblock {\em Advances in Services Marketing and Management}, 5:127--151.

\bibitem[\protect\citename{Jones}2002]{Jones02}
Steven~Swann Jones.
\newblock 2002.
\newblock {\em The Fairy Tale: The Magic Mirror of the Imagination}.
\newblock Routledge.

\bibitem[\protect\citename{Kast}1993]{Kast93}
Verena Kast.
\newblock 1993.
\newblock {\em Through Emotions to Maturity: Psychological Readings of Fairy
  Tales}.
\newblock Fromm Intl.

\bibitem[\protect\citename{Lebert}2009]{Lebert09}
Marie Lebert.
\newblock 2009.
\newblock {\em Project {G}utenberg (1971--2009)}.
\newblock Benediction Classics.

\bibitem[\protect\citename{Lehrer}1974]{Lehrer74}
Adrienne Lehrer.
\newblock 1974.
\newblock {\em Semantic fields and lexical structure}.
\newblock North-Holland, American Elsevier, Amsterdam, NY.

\bibitem[\protect\citename{Lobo and Martins~de Matos}2010]{Lobo10}
Paula~Vaz Lobo and David Martins~de Matos.
\newblock 2010.
\newblock Fairy tale corpus organization using latent semantic mapping and an
  item-to-item top-n recommendation algorithm.
\newblock In {\em Language Resources and Evaluation Conference - LREC 2010,
  European Language Resources Association (ELRA)}, Malta.

\bibitem[\protect\citename{Mannix}1992]{Mannix92}
Patrick Mannix.
\newblock 1992.
\newblock {\em The rhetoric of antinuclear fiction: Persuasive strategies in
  novels and films}.
\newblock Bucknell University Press, Associated University Presses, London.

\bibitem[\protect\citename{Michel \bgroup et al.\egroup }2011a]{GBC}
Jean-Baptiste Michel, Yuan~K. Shen, Aviva~P. Aiden, Adrian Veres, Matthew~K.
  Gray, The Google~Books Team, Joseph~P. Pickett, Dale Hoiberg, Dan Clancy,
  Peter Norvig, Jon Orwant, Steven Pinker, Martin~A. Nowak, and Erez~L. Aiden.
\newblock 2011a.
\newblock {Quantitative Analysis of Culture Using Millions of Digitized Books}.
\newblock {\em Science}, 331(6014):176--182.

\bibitem[\protect\citename{Michel \bgroup et al.\egroup }2011b]{Michel11}
Jean-Baptiste Michel, Yuan~Kui Shen, Aviva~Presser Aiden, Adrian Veres,
  Matthew~K. Gray, The Google~Books Team, Joseph~P. Pickett, Dale Hoiberg, Dan
  Clancy, Peter Norvig, Jon Orwant, Steven Pinker, Martin~A. Nowak, and
  Erez~Lieberman Aiden.
\newblock 2011b.
\newblock Quantitative analysis of culture using millions of digitized books.
\newblock {\em Science}, 331(6014):176--182.

\bibitem[\protect\citename{Mohammad and Turney}2010]{MohammadT10}
Saif~M. Mohammad and Peter~D. Turney.
\newblock 2010.
\newblock Emotions evoked by common words and phrases: Using mechanical turk to
  create an emotion lexicon.
\newblock In {\em Proceedings of the NAACL-HLT 2010 Workshop on Computational
  Approaches to Analysis and Generation of Emotion in Text}, LA, California.

\bibitem[\protect\citename{Mohammad and Yang}2011]{MohammadY11}
Saif~M. Mohammad and Tony~(Wenda) Yang.
\newblock 2011.
\newblock Tracking sentiment in mail:\\ how genders differ on emotional axes.
\newblock In {\em Proceedings of the ACL 2011 Workshop on Computational
  Approaches to Subjectivity and Sentiment Analysis (WASSA)}, Portland, OR,
  USA.

\bibitem[\protect\citename{Mohammad \bgroup et al.\egroup }2009]{MohammadDD09}
Saif~M. Mohammad, Cody Dunne, and Bonnie Dorr.
\newblock 2009.
\newblock Generating high-coverage semantic orientation lexicons from overtly
  marked words and a thesaurus.
\newblock In {\em Proceedings of Empirical Methods in Natural Language
  Processing (EMNLP-2009)}, pages 599--608, Singapore.

\bibitem[\protect\citename{Mohammad}2011]{Mohammad11a}
Saif~M. Mohammad.
\newblock 2011.
\newblock Even the abstract have colour: Consensus in word–colour
  associations.
\newblock In {\em Proceedings of the 49th Annual Meeting of the Association for
  Computational Linguistics: Human Language Technologies}, Portland, OR, USA.

\bibitem[\protect\citename{Orenstein}2003]{Orenstein03}
Catherine Orenstein.
\newblock 2003.
\newblock {\em Little Red Riding Hood Uncloaked: Sex, Morality, And The
  Evolution Of A Fairy Tale}.
\newblock Basic Books.

\bibitem[\protect\citename{Pang and Lee}2008]{PangL08}
Bo~Pang and Lillian Lee.
\newblock 2008.
\newblock Opinion mining and sentiment analysis.
\newblock {\em Foundations and Trends in Information Retrieval},
  2(1--2):1--135.

\bibitem[\protect\citename{Plutchik}1980]{Plutchik80}
Robert Plutchik.
\newblock 1980.
\newblock A general psychoevolutionary theory of emotion.
\newblock {\em Emotion: Theory, research, and experience}, 1(3):3--33.

\bibitem[\protect\citename{Stone \bgroup et al.\egroup }1966]{Stone66}
Philip Stone, Dexter~C. Dunphy, Marshall~S. Smith, Daniel~M. Ogilvie, and
  associates.
\newblock 1966.
\newblock {\em The General Inquirer: A Computer Approach to Content Analysis}.
\newblock The MIT Press.

\bibitem[\protect\citename{Strapparava and Valitutti}2004]{StrapparavaV04}
Carlo Strapparava and Alessandro Valitutti.
\newblock 2004.
\newblock Wordnet-{A}ffect: {A}n affective extension of {W}ord{N}et.
\newblock In {\em Proceedings of the 4th International Conference on Language
  Resources and Evaluation (LREC-2004)}, pages 1083--1086, Lisbon, Portugal.

\bibitem[\protect\citename{Turney and Littman}2003]{Turney03}
Peter Turney and Michael Littman.
\newblock 2003.
\newblock Measuring praise and criticism: Inference of semantic orientation
  from association.
\newblock {\em ACM Transactions on Information Systems (TOIS)}, 21(4):315--346.

\bibitem[\protect\citename{Yarowsky}1992]{Yarowsky92}
David Yarowsky.
\newblock 1992.
\newblock Word-sense disambiguation using statistical models of {R}oget's
  categories trained on large corpora.
\newblock In {\em Proceedings of the 14th International Conference on
  Computational Linguistics (COLING-92)}, pages 454--460, Nantes, France.

\end{thebibliography}


\end{document}